\documentclass[sn-basic,Numbered]{sn-jnl}%

\usepackage{graphicx}%
\usepackage{multirow}%
\usepackage{amsmath,amssymb,amsfonts}%
\usepackage{amsthm}%
\usepackage{mathrsfs}%
\usepackage[title]{appendix}%
\usepackage{xcolor}%
\usepackage{textcomp}%
\usepackage{manyfoot}%
\usepackage{booktabs}%
\usepackage{algorithm}%
\usepackage{algorithmicx}%
\usepackage{algpseudocode}%
\usepackage{listings}%
\usepackage{adjustbox}
\usepackage{multirow}
\usepackage{booktabs}
\usepackage{setspace}
\usepackage{tabularx}

\theoremstyle{thmstyleone}%
\theoremstyle{thmstyletwo}%

\theoremstyle{thmstylethree}%

\begin{document}

\title[Article Title]{Deep Video Representation Learning: a Survey}

\author[1]{\fnm{Elham} \sur{Ravanbakhsh}}\email{eravan1@lsu.edu}

\author[2]{\fnm{Yongqing} \sur{Liang}}\email{lyq@tamu.edu}

\author[1]{\fnm{J.} \sur{Ramanujam}}\email{jxr@cct.lsu.edu}

\author*[3,2]{\fnm{Xin} \sur{Li}}\email{xinli@tamu.edu}

\affil[1]{\orgdiv{Division of Electrical $\&$ Computer Engineering and Center for Computation $\&$ Technology}, \orgname{Louisiana State University}, \orgaddress{\city{Baton Rouge}, \postcode{70803}, \state{LA}, \country{USA}}}

\affil[2]{\orgdiv{Department of Computer Science and Engineering}, \orgname{Texas A$\&$M University}, \orgaddress{\city{College Station}, \postcode{77843}, \state{TX}, \country{USA}}}

\affil[3]{\orgdiv{Section of Visual Computing and Interactive Media}, \orgname{Texas A$\&$M University}, \orgaddress{\city{College Station}, \postcode{77843}, \state{TX}, \country{USA}}}


\abstract{This paper provides a review on \emph{representation learning for videos}. We classify recent spatio-temporal feature learning methods for sequential visual data and compare their pros and cons for general video analysis. 
Building effective features for videos is a fundamental problem in computer vision tasks involving video analysis and understanding. Existing features can be generally categorized into spatial and temporal features. Their effectiveness under variations of illumination, occlusion, view and background are discussed. Finally, we discuss the remaining challenges in existing deep video representation learning studies.}


\maketitle

\section{Introduction}\label{sec1}
The enormous influence of media and social networking has led to an avalanche of videos uploaded on the internet every day. To effectively analyze and use the uploaded video data, it is important to construct feature representations for videos. 
Unlike the analysis and understanding of images, the manual modeling of video features is often a laborious task. Therefore, there is a need for techniques that can automatically extract compact yet descriptive features. 

With the recent advances in artificial intelligence and computer vision, deep neural networks have achieved significant success in feature modeling. 
These techniques have led to a great breakthrough in practical video analysis applications such as tracking~\cite{liang2018tracking,xu2017lie}, action recognition~\cite{yan2018spatial}, action prediction~\cite{liang2019peeking}, and person re-identification~\cite{hou2019vrstc}. 
To design a deep learning pipeline for these applications, extracting video features is often the first step and it plays a critical role in subsequent video processing or analysis. 
Developing deep learning pipelines to extract effective features for a given video is referred to as \emph{deep video representation learning}.

The characteristics of videos are often encoded by \emph{spatial features} and \emph{temporal features}. Spatial features encode geometric structures, spatial contents, or positional information in image frames; whereas, temporal features capture the movements, deformation, and various relations between frames in the time domain. 
Depending on the target applications, an algorithm should be able to extract either spatial or temporal features, preferably, both. Some applications also require the decoupling of spatial and temporal information from the extracted features so that some specific characteristics can be more effectively modeled. 

Learning robust representation for videos faces several \emph{major challenges} such as 
\begin{itemize}
    \item occlusion: objects of interests might be partly occluded;
    \item illumination: videos might be taken under various lighting conditions or/and from changing view angles;
    \item view and background variations: foreground objects and background scenes can be moving.
\end{itemize} 
Therefore, we evaluate the performance of representation learning algorithms using \emph{robustness} and \emph{accuracy} under these scenarios. 
Robustness of different algorithms is evaluated under these four challenges: occlusion, view, illumination, and background change. 
As for their accuracy, since different applications use different metrics, we adopt accuracy metrics from representative tasks of action recognition and video segmentation, in which more expressive features generally lead to better accuracy. 


\textbf{Comparison with Existing Surveys.} 
Representation learning from images is a classic problem in computer vision, and it has been widely studied to facilitate various image analysis and understanding tasks. Many survey papers have been published to address this problem. But most of these representation learning studies focused on features of static images~\cite{ma2021image,kapoor2021state,piasco2018survey,minaee2021image}. 
As summarized in Table~\ref{Tab:surveys}, while multiple components of these studies are closely related to video representation, a systematic survey on video features is missing. Some recent surveys ~\cite{hamilton2017representation,li2018survey,jing2020self} discussed video representation learning, but most of these have focused on a specific type of learning or method. 
A few other survey papers discussed video processing tasks that involve video representation learning~\cite{cuevas2020techniques, de2021monocular, hussain2021comprehensive, WANG2022108220,karbalaie2022event}; however, their focuses were mostly on discussing how the developed pipelines perform on the targeted task(s). 
There is a lack of a survey of representation learning in a general setting and one that investigates the role of each feature on its embedding regardless of its specific task. 
We believe our survey provides an insightful analysis on how to construct effective video features/representations, particularly when confronting with aforementioned challenges.  

\begin{enumerate}
    \item We provide a comprehensive survey of deep video representation learning.
    \item We compare different types of representation learning algorithms in terms of accuracy and robustness in 
    various practical challenging scenes, and provide some observations/suggestions in adopting suitable features for different video processing and analysis tasks. 
\end{enumerate}

\begin{table}[h!tb]
\caption{\label{Tab:surveys}Current surveys related to image and video feature learning. 
Unlike existing surveys that studied representation learning for specific targeting tasks, our survey discusses pros and cons of different features for general scope.} 
\centering

\begin{tabular}{|p{0.08\textwidth} | p{0.05\textwidth} | p{0.05\textwidth} |p{0.05\textwidth} |p{0.1\textwidth} | p{0.47\textwidth}|} \hline
 Reference & Year & Image & Video & General scope & Application  \\ \hline
 \cite{piasco2018survey} & 2018 & \checkmark & - & $\times$ & Visual-based localization\\
 \cite{ma2021image} & 2021 & \checkmark & - & $\times$ & Image matching\\ 
 \cite{minaee2021image} & 2021 & \checkmark & - & $\times$ & Semantic and instance segmentation\\ 
 \cite{kapoor2021state} & 2021 & \checkmark & - & $\times$ & Content-based image retrieval \\ 
 \cite{hamilton2017representation} & 2017 & \checkmark & \checkmark & $\times$ & Representation learning on graphs\\ 
 \cite{li2018survey} & 2018 & \checkmark & \checkmark & $\times$ & Multi modal learning \\ 
 \cite{jing2020self} & 2020 & \checkmark & \checkmark & $\times$ & Self supervised feature learning\\  
 \cite{song2021human} & 2021 & \checkmark & \checkmark & $\times$ & 2D and 3D pose estimation \\ 
 \cite{de2021monocular} & 2021 & \checkmark & \checkmark & $\times$ & Multi person pose estimation \\ 
 \cite{cuevas2020techniques} & 2020 & - & \checkmark & $\times$ & Soccer video analysis\\
 \cite{hussain2021comprehensive} & 2021 & - & \checkmark & $\times$ & Multi-view video summarization\\ 
 \cite{karbalaie2022event} & 2022 & - & \checkmark & $\times$ & Event detection in surveillance videos\\ 
  \cite{WANG2022108220} & 2022 & - & \checkmark & $\times$ & Pedestrian attribute recognition\\ 
 \textbf{Ours} & \textbf{2023} & \checkmark & \checkmark & \checkmark & \textbf{Action recognition + video segmentation} \\
  \hline
\end{tabular}
\end{table}

\textbf{Organization.} We present the classification of deep features for videos in Section~\ref{Sec:Features}, and then compare these features in action recognition and video segmentation application in Section~\ref{Sec:Applications}. We conclude the paper by discussing remaining challenges and future directions in Section~\ref{Sec:Conclusion}.

\section{Classification of Deep Video Features}
\label{Sec:Features}

Two main aspects of video data are often considered in video processing and analysis tasks: (1) how to encode spatial structures or contents in visual data; and (2) how to model temporal coherency or changes among frames. 
Appearance information and geometry structure of the scene or objects are considered as \textbf{spatial information}. 
In representation learning, capturing spatial relation is important in understanding the visual concept of the video.
Based on spatial information extracted, we generally divide features into \textbf{dense} features and \textbf{sparse} features. \textbf{Spatially dense} features are contextual data often defined using pixel intensities of the input. Typical and widely used dense features can come from RGB images and their variants, such as RGBD images. 
\textbf{Spatially sparse} features are often defined on a smaller set of entities such as keypoints, subpatches, or other graph structures. Widely used sparse features include those defined on divided patches or structural graphs (e.g., media axes for general shapes, skeletal structures of humans/animals). 

\begin{figure}[h!tb]
\centering
\includegraphics[width=200pt,keepaspectratio]{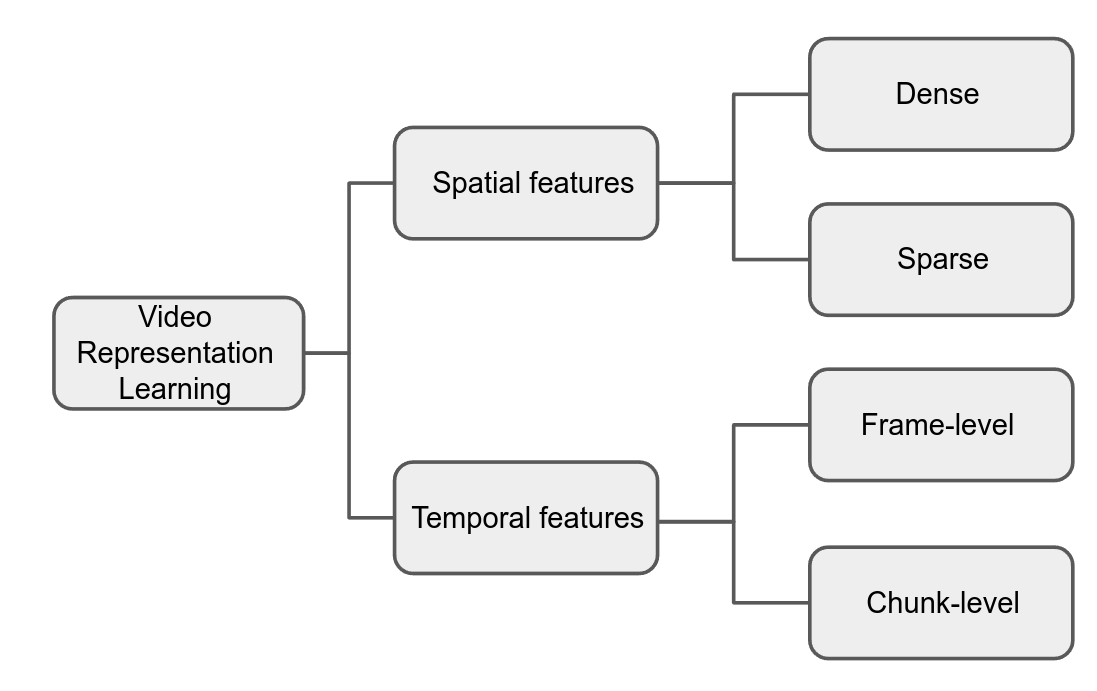}

\caption{\label{tab:organization} Classification of deep video representation learning schemes.}
\end{figure}

\begin{table}[h!tb]
\centering
\caption{\label{Tab:FeatureClassificationProsAndCons}Pros and cons of different types of features. Scene noise includes occlusion, illumination, background and viewpoint variations.}
\begin{tabular}{|p{0.1\textwidth} | p{0.4\textwidth} | p{0.4\textwidth}|}   
\hline
 Features & Pros & Cons \\ \hline
 Dense & Good in capturing appearance information & Sensitive to scene noise\\ 
  & & High intra-class variations \\
  & & High computational cost \\
 \hline
 Sparse & Robust against background and illumination change & Weak in capturing appearance information\\
  & Low intra-class variations &  \\
  & Low computational cost &  \\
 \hline
 Frame-level & Low computational cost & Weak in co-occurrence representation learning \\ 
 \hline
 Chunk-level & Good in co-occurrence representation learning & High computational cost \\ 
 \hline
\end{tabular}
\end{table}

The second aspect is \textbf{temporal information} which plays an important role in video representation learning, and is a key difference between image features and video features. 
To effectively understand the concept of a video, temporal information or temporal coherence across different frames plays a critical role. 
Based on how the temporal information is modeled, we divide temporal features into two categories: \textbf{frame-level} and \textbf{chunk-level} features. The former extracts features from each frame and constructs a sequence of signatures; while the later one encodes spatio-temporal features of a chunk into one signature. 

We illustrate our classification in Fig.~\ref{tab:organization}.  
In the following subsections~\ref{Spatially dense features} to \ref{Chunk-level features}, we classify video features based on how their spatial and temporal information is modeled, and discuss their pros and cons, as summarized in Table~\ref{Tab:FeatureClassificationProsAndCons}.
We will discuss how different designs affect the features' general robustness under different scenarios.  In terms of feature accuracy, it is more application-dependent, and will be discussed in the application section.

\subsection{Spatially dense features} 
\label{Spatially dense features}
Spatially dense features contain rich information mostly defined by using direct pixel intensities of the input in a structured order. RGB video is the most common data type in dense features. It contains contextual data about appearance, objects and background. People often use a 2D Convolutional Neural Network (CNN) as \emph{a standard architecture} for extracting spatial information from dense features.

\textbf{CNN-based Standard Architectures.}
With the emergence of deep learning, CNNs have become the most common method for feature modeling due to the strong modeling capability and superior performance of deep learning-based methods. Various CNN architectures including VGGNet~\cite{simonyan2014very} and ResNet-50~\cite{he2016deep} have been used for spatial representation learning \cite{liu2020video, zhao2019spatiotemporal,eun2020learning}.
These models provide a high-level spatial representation of video frames. 
Some approaches also use object detection or segmentation algorithms to extract local regions of interest in a frame to better exploit the correlation of different regions and reduce the chance of encoding redundant information \cite{bendre2022natural}. 
For example, in \cite{liang2019peeking}, an interaction module is proposed to model interactions between a (human) object and it's surroundings. Mask-RCNN~\cite{he2017mask} is used for semantic segmentation, then these masked features are given to a 2-layer convolutional network for further feature learning. 

\textbf{Extra modules for better robustness.} 
Effective video features should be robust against occlusion, view and background change. Therefore, built upon the above \emph{standard architectures}, people also add extra modules to improve feature robustness. 
Recent approaches adopted three general types of additional modules: \emph{part information}, \emph{additional input information}, and \emph{attention mechanism}. 
In the following, we will elaborate that given a \emph{standard architecture} X, adding one or multiple modules onto X could improve the pipeline's robustness under different scenarios.

\paragraph{Part information}
Typical video representation learning methods do not take into account the effect of partial occlusion. But with partial occlusion in the video, learned features are often corrupted due to the inclusion of irrelevant surrounding objects, and consequently, cause dramatic performance degradation. 
A conventional strategy is to train ensemble models for various occlusion patterns \cite{tian2015deep, ouyang2012discriminative, chen2015parsing}, and construct a part pool that covers different scales of object parts, and then automatically choose important parts to overcome occlusions. 
A main limitation of this strategy is that it is more expensive in both training and testing phases. 
Another strategy is to integrate a set of part/occlusion-specific detectors in a joint framework that learns partial occlusions \cite{zhou2017multi, ouyang2013joint}. 
In \cite{zhou2017multi}, a set of part detectors are applied and the correlation among them is exploited to improve their performance. Each part detector generates a score for a candidate region and the final score is computed based on the average among some top scores. The main issue with such part-specific detectors is that they are not able to cover all the occlusion patterns comprehensively and need to be designed based on some pre-assumptions \cite{xu2017jointly}. For example, for pedestrian detection, these part detectors are designed with the prior knowledge that pedestrians are usually occluded from the bottom, left, and right. However, in practice, occlusion patterns can be irregular, which affects the feature's performance. 

\paragraph{Additional Information} Some studies use additional information to compensate for the view and the illumination variance of dense features. For example, RGB-D cameras are often used to include depth information which makes the representation less sensitive to illumination and view variations \cite{hu2015jointly,liu2017viewpoint}. Thermal images together with color images are also used to improve robustness when suitable light source is not available~\cite{kniaz2018thermalgan}. Several other works use non-vision information as a complementary input. For example, in \cite{gao2020listen}, audio signals and in \cite{liu2021semantics} signals from wearable sensors are added to improve robustness of dense features.

\paragraph{Attention Mechanism} 
In most scenarios, key objects/regions are just part of the whole spatial image. Being able to use local spatial attention to guide the system to focus on important (foreground) object and ignore irrelevant (background) noise is desirable. 
Many recent studies developed attention mechanisms to help feature learning models concentrate on important regions in the spatial dimension. For example, in \cite{xu2017jointly}, an attentive spatial pooling is used instead of max-pooling which computes similarity scores between features to compute attention vectors in the spatial dimension. This method allows model to be more attentive on region of interests in image level.
In \cite{li2021diverse}, a self-attention mechanism is adopted to generate a context-aware feature map. Then the similarity between context-aware feature maps and a set of learnable part prototypes are calculated and used as spatial attention maps. To identify the object of interest and focus on visible parts, some models use spatially sparse features in spatial attention networks to construct a robust representation. In \cite{zhang2018occluded, xu2018attention}, sparse features aid RGB frames to estimate the attention map and visibility scores to handle various occlusion patterns and generate view-invariant representation. 

\paragraph{Summary of Extra Modules} 
Although dense features can encode rich contextual information,  their representation can sometimes be sensitive to occlusion, view, illumination, and background variance. 
Given a standard architecture $X$ that extracts dense features, 
various recent studies exploited adding extra modules onto $X$ to enhance the robustness of spatial feature modeling. Table~\ref{Tab:rgb data} summarizes these strategy discussed above.  $X$ + part information is often effective in enhancing the model's robustness against partial occlusion. 
$X$ + additional input information (e.g., depth info) can help the model enhance its performance under illumination and view changes.
Many recent studies incorporate attention mechanism into the standard architecture, such an attention strategy often help 
differentiate foreground objects of interest and the background noise, and consequently, helps the model perform better towards view, occlusion, and background variance.

\begin{table}[h!tbp]
\centering
\caption{\label{Tab:rgb data} Robustness of different models using Dense features. A standard architecture is called X. Its limitations against robustness challenges  can be addressed/alleviated by adding extra modules.}
\begin{tabular}{|p{0.3\textwidth} | p{0.1\textwidth} | p{0.13\textwidth}| p{0.15\textwidth}| p{0.15\textwidth}|}
\hline
  Model & View & Occlusion  & Background & Illumination \\
  \hline
  X & - & - & - & - \\
  X + part information  & - & \checkmark & - & - \\ 
  X + additional information  & \checkmark & - & - & \checkmark \\
  X + attention & \checkmark & \checkmark & \checkmark & - \\
 \hline
\end{tabular}
\end{table}

\subsection{Spatially sparse features}
\label{Spatially sparse features}

Spatially sparse features represent information with a sparse set of feature points that are able to describe the geometry of original video frames. There are several advantages of using sparse features. First, they describe frames with a sparse set of features that leads to a low computational cost. Second, sparse features suffer less intra-class variances compared to dense features and are more robust to the change of conditions such as appearance, illumination, and backgrounds. However, sparse features lack appearance information which in some scenarios is essential for feature modeling. 
Three types of \emph{standard architectures} are often adapted to extract sparse features: \emph{RNN, CNN, and GNN/GCN pipelines.}

\textbf{RNN-based Standard Architectures.} 
Due to the recurrent nature of videos, one strategy is to use RNN-based architectures for sparse representation learning. 
Spatial feature arrangements determine the proper architecture for spatial extraction. If the inherent spatial relations between feature points can be arranged in a sequence of vectors or grids, RNNs are effective for spatial modeling \cite{veeriah2015differential,du2015hierarchical}. 

\emph{Pros and Cons of RNN-based Standard Architectures} 
RNN-based methods are good in dealing with sequential data, but they are not very effective in spatial modeling. Therefore, their general performance is not as good as CNN models~\cite{liu2017global}. In several works, RNN-based models take feature maps from CNNs as their input rather than using raw input frames~\cite{xu2019temporal,eun2020learning}.

\textbf{CNN-based Standard Architectures.} 
There are many studies that deployed CNNs for sparse representation learning. However, defining the relations between unstructured feature points is challenging. CNNs take their input in a form of an image. 
To satisfy the need of CNNs’ input, some researches model the sparse features into multiple 2D pseudo-images \cite{liu2017skeleton, liu2017enhanced, li2017skeletonn}. 
For example, in \cite{liu2017skeleton}, sparse features are arranged into an image containing the feature points in one dimension and frames in another one, then a 7-layer CNN is applied to extract features. In \cite{wang2016action}, the frame dynamics are mapped into textured images, then a CNN model is used to extract information. 
Some works use a group of feature points to construct a hierarchy among features.
In~\cite{herzig2022object}, the video frame is divided into multiple grids, then a convolutional feature descriptor is run for each cell.

\emph{Pros and Cons of CNN-based Standard Architectures} CNN-based architectures are effective in extracting local features and exploring discriminative patterns in data. However, their major drawback is that they are designed for image-based input and primarily rely on spatial dependencies between the neighboring points. For some sparse features that contain unstructured design, CNNs cannot perform very well. Additionally, CNNs have trouble with wildly sparse data as they heavily rely on spatial relations of points to learn. In that case, RNNs perform better. 

\textbf{GNN/GCN Standard Architectures. }
Modeling sparse features in a vector or an image may corrupt the spatial relations or add false connections between feature points that their relation is not strong enough. Some of these irregular features are intrinsically structured as a graph and cannot be modeled in a vector, 2D or 3D grid. To address this issue, some researches utilize graph structure, such as graph convolutional neural networks (GCNNs) ~\cite{yan2018spatial, li2019spatio, li2019actional} or GNNs \cite{hao2021hypergraph} to build feature extraction architectures.

Some people use spatio-temporal GCNs by arranging feature points as an indirect graph with points as nodes and their relations as edges. Nodes within one frame are connected based on the spatial relations of the features and represented by an adjacency matrix in spatial dimension. Nodes in the temporal dimension are related through the relations of corresponding nodes in consecutive frames \cite{yan2018spatial}. A concern with spatio-temporal GCNs is that modeling spatial features just according to natural connection of nodes might lose the potential dependency of disconnected joints. To solve this issue, in \cite{li2019actional, shi2019skeleton}, a model is introduced to adaptively learn and update the topology of graph for different layers and samples. 
In \cite{li2019actional}, a framework is proposed to capture richer dependencies among points and neighbors. They used an encoder-decoder structure to model dependencies between far-apart points. When modeling sparse features, defining connections between different feature nodes is challenging. For example, spatio-temporal GCN models the features as an indirect graph, which may not fully express the direction and position of points. \cite{shi2019skeleton} used a directed acyclic graph to represent sparse data. Then a directed graph neural network is used to encode the constructed directed graph, which can propagate the information to adjacent nodes and edges and update their associated information in each layer.

\emph{Pros and cons of GNN/GCN Standard Architectures}
GNN/GCNs are helpful in dealing unstructured data that have underlying graph structures and are non-Euclidean. The non-regularity of data structures in most sparse features have led to superior performance of graph neural networks over CNN or RNN architectures. 
But if sparse features can be formulated as 2D grids or vectors, then graph-based models are not as efficient as CNN models.

\medskip
\textbf{Extra Modules for Better Robustness.} 
The representation of spatially sparse features is robust to background and illumination change. 
To overcome other robustness issues, some studies add extra modules on the standard architecture. Standard architecture X could be a CNN, RNN or GNN/GCN model. Generally, two extra modules, \emph{transformation matrix} and \emph{attention}, are often added to enhance feature robustness.

\paragraph{Transformation matrix} Change of camera view points can change the relative position of feature points. Hence, constructing a view-invariant representation remains still a challenge in sparse feature modeling. To address this problem, several researches developed a transformation method to transform a set of feature points to a standard structure \cite{huang2017deep, li2020learning, liu2017enhanced}. In \cite{li2020learning}, a rotor-based view transformation method is proposed to re-position the original features to a standard frontal system. After transformation, a spatio-temporal model is applied to construct the shape and motion of each part. In \cite{liu2017enhanced}, a sequence-based transformation is applied on the features to map them to a standard form and make a new view-invariant sequence.

\paragraph{Attention mechanism} Attention aids model in confronting with partial occlusion and view variations. Similar to dense features, sparse feature modeling in the presence of occlusion is still challenging. Some approaches deploy spatial attention mechanism to focus on the points of interest and predict occluded parts by the help of visible feature points in the adjacent frames. In \cite{hou2019vrstc}, a spatial attention generator is proposed to predict occluded parts. The generator is an autoencoder that predicts the content of occluded part conditioned on the visible parts of the current frame. Some other studies use heatmaps as attention mechanism to focus on informative feature points. In \cite{Fabbri_2018_ECCV}, heatmaps for the visible and occluded part are generated. Then, using the heatmaps occluded parts are predicted along both spatial and temporal dimension. In \cite{song2019richly}, class activation maps are used as a mask matrix to force  network has to learn features from currently inactivated points. 
To achieve a view-invariant representation, some models propose transferring attention from reference view to arbitrary views. For example, in \cite{ji2021view}, attention maps are produced to transfer attention from a reference view to arbitrary views. This helps learn effective attention to crucial feature points. In \cite{wang2020learning}, attention directly operates on network parameters rather on input features. This allows spatial interactional contexts to be explicitly captured in a unified way.

\emph{Summary.} Table \ref{Tab:sparse} shows the robustness of standard architecture X and extra modules. X could be RNN, CNN or GNN/GCN architecture and is robust against background and illumination changes. While pros and cons of X depend on the target task, generally RNN is better suited to more sparse data while CNN deals better with denser feature points. Typically if the nature of the structure of data is in a grid format, CNN is more effective in extracting spatial features. On the other hand, if data have a graph scheme, then GNN/GCN performs better. X + transformation matrix is robust against view variations by transforming a set of feature points to a standard view. X + attention is robust against occlusion and view variations by predicting occluded parts from visible points and transfer attention from one view to another, respectively.

\begin{table}[h!tb]
\centering
\caption{\label{Tab:sparse} Robustness of different models using spatially sparse features. A standard architecture is called X.}
\begin{tabular}{|p{0.3\textwidth} | p{0.1\textwidth} | p{0.1\textwidth}| p{0.15\textwidth}| p{0.15\textwidth}|}
 \hline
  model & view & occlusion & background & illumination \\
  \hline
  X & - & - & \checkmark & \checkmark \\
  X + transformation matrix & \checkmark & - & \checkmark & \checkmark \\
  X + attention & \checkmark & \checkmark & \checkmark & \checkmark \\
\hline
\end{tabular}
\end{table}

\subsection{Frame-level features}
\label{Frame-level features}

Frame-level features are a sequence of signatures that each describes a frame individually. Given a video clip, an frame-level feature model often processes spatial information frame by frame and  encodes the temporal relationship between frames. People adopt different strategies, such as \emph{Optical flow}, \emph{CNN-based architectures}, \emph{RNN-based architectures}, and \emph{Attention mechanisms} to extract frame level features. 

\textbf{Optical Flow.} Optical flow is a feature containing motion information of consecutive frames that is useful for describing video dynamics \cite{horn1981determining}. It is computed by removing the non-moving scene that generates a background invariant representation compared to the original frames. As optical flow is computed frame by frame we categorize it as a frame-level feature. Most studies use optical flow as a temporal data along with their original spatial features to capture movements. 
Studies in \cite{carreira2017quo,wang2016temporal,wang2015towards}, showed that using optical flow and RGB frames achieves a superior performance in modeling videos than only using RGB frames. In \cite{wang2016temporal}, optical flow of consecutive 10 frames and RGB are fed to a CNN. The convolutional filters compute derivatives of the optical flow and learn motion dynamics w.r.t the image location. Although to some extent, extracting optical flows as an additional information helps models be more view-independent, robust to occlusion and cluttered background, it often cannot capture relatively long term dependencies. In addition to optical flow, there are other features that contain motion information. For example, in \cite{choutas2018potion}, a new motion representation called Potion is proposed that provides all the dynamic of an instance throughout the video clip in one image. Motion cues are represented by different colors which shows the relative time of the frame in the video clip. Using Potion along with RGB frames and optical flows aid model to encode relatively longer dependencies. In \cite{zong2021motion}, in addition to optical flow, motion saliency \cite{chen2016motion} is calculated from consecutive video frames.  Pre-computing motion information including optical flow is time-consuming and storage-demanding. Also, they cannot learn global dependencies among frames.

\textbf{CNN-based Architectures.} As CNNs have been widely used in the image tasks, some studies \cite{he2019stnet, lin2019tsm, wang2021tdn}
adopted 2D CNNs to model the video clips. However, 2D CNNs need a fusion module to concatenate temporal cues. For example, in \cite{Karpathy_2014_CVPR}, late fusion is adopted which fuses information from two different CNNs in the first fully connected layer. In~\cite{wu2022multi}, the authors use 2D CNN networks to encode each frame into feature maps then concatenate them as a long vector as the video-level feature descriptor to holistic understanding of the video. However, using a simple fusion or average lacks the learning capability and doesn't contain useful time-related features. In \cite{lin2019tsm}, the channels along the time dimension is shifted to improve the performance of temporal modeling with 2D CNNs. In \cite{yan2022gl}, consecutive frames are aggregated with adaptive weights and then fed to convolutional networks.

Recent papers extend the single pass feature encoder to the Siamese structure for time-related feature learning. 
The Siamese pipeline takes two frames as inputs and uses CNN encoders to extract feature maps.
In~\cite{chen2018blazingly,hu2018videomatch}, their feature encoders shared the same weights to extract feature maps from the input images.
Then they compare the similarity between the feature maps to build the inter-frame features.
In~\cite{oh2019video,liang2020video,li2022recurrent}, they use two independent encoders to extract features maps. 
They encode the features into two parts, one is for similarity comparison, the other stores semantic information of frames.
These video features are more useful in the down-stream tasks.

\textbf{RNN-based Architectures.} Due to the sequential nature of videos and the ability of memorizing the temporal relations in RNN-based architectures, some studies used these models for encoding motions in a video. Some studies used parallel stream architecture which one stream is responsible for extracting spatial information and the other stream is responsible for extracting temporal data. In \cite{wang2017modeling}, a two-stream architecture is proposed where the first stream is responsible for learning spatial dependency and the second for learning the temporal dynamics. The two streams are then aggregated with each other to represent data. As RNNs are not effective in learning long-term temporal dependencies, most models adopt LSTM models. In \cite{zhao2019spatiotemporal,eun2020learning}, LSTM is fed with the spatial representation of each frame at each time step. Learning process at each time step is based not only on the observations at that time step, but also on the previous hidden states that provide temporal context for the video. Some other versions of Recurrent-based netwroks were also explored. For example, in \cite{eun2020learning}, an extended GRU is proposed to model temporal relations by using current and old information. GRU requires less storage and performs faster.

\textbf{Attention Mechanism.}
As not all frames are informative, several studies used temporal attention to discriminate and select key frames by assigning weights to them. In \cite{chen2018video} a scaled dot product attention module is adopted that assigns weights to frame features according to their importance. In \cite{xu2017jointly}, all time steps resulting from RNN are combined by an attentive temporal pooling to compute an attentive score in temporal dimension to weight frames based on their goodness. In \cite{zhang2019scan}, a non-parametric self and collaborative attention network is proposed to efficiently calculate the correlation weights to align discriminative frames. In \cite{lu2023transflow}, a transformer module is adopted that iteratively chooses one frame
as query and the rest as key features to compute the temporal attention.

\begin{table}[!tb]
\centering
\caption{\label{Tab:frame-level}Pros and cons of various architectures used for modeling frame-level features}
\begin{tabular}{|p{0.085\textwidth} | p{0.42\textwidth} | p{0.4\textwidth}|}   
\hline
 \textbf{Models} & \textbf{Pros} & \textbf{Cons} \\ \hline
 Optical  & Effective for simple and local dynamics & Ineffective for long and complex dynamics\\ 
 flow &  & Hard to handle temporal scale variance\\
  &  & Expensive computation and storage \\
 \hline
 2D CNN & Effective for simple and local dynamics & Ineffective for long and complex dynamics\\
 & & hard to handle temporal scale variance\\
 \hline
 RNN & Effective for complex dynamics & Hard to handle temporal scale variance \\ 
 \hline
 Attention & Effective for long and complex dynamics & Expensive computation  \\
 & Effectively handle temporal scale variance  \\
 \hline
\end{tabular}
\end{table}

\emph{Summary.} As shown in table \ref{Tab:FeatureClassificationProsAndCons}, while frame-level features have lower computational costs compared to chunk-level features, they decouple spatial and temporal dimensions which leads to lack of co-occurrence representation learning. People adopt different methods for learning frame-level dynamics in a video as shown in table \ref{Tab:frame-level}. Optical flow and CNN-based architectures assist network to capture short and simple cues. However, they suffer from the lack of memory and therefore they are not suitable for capturing long term dependencies. Additionally, optical flow is computationally expensive and storage demanding. RNN-based architectures, particularly LSTMs, are able to encode complex dynamics, however they treat each video clip equally and are invariant to inherent temporal diversities. Attention selects informative features and is suited for distinguishing complex tasks.    


\subsection{Chunk-level features} 
\label{Chunk-level features}
\begin{table}[h!tb]
\caption{\label{tab:chunk-level}Pros and cons of various dynamics used for modeling chunk-level features}
\begin{tabular}{|p{0.11\textwidth} | p{0.47\textwidth} | p{0.3\textwidth}|}   
\hline
 \textbf{Models} & \textbf{Pros} & \textbf{Cons} \\ \hline
 3D CNN & effective for coarse level dynamics & ineffective for fine dynamics\\ 
  &  & expensive computation \\
 \hline
 Attention & effective for fine and coarse level dynamics & expensive computation \\
 & effectively handle temporal scale variance & \\
 \hline
\end{tabular}

\end{table}

While frame-level features separate spatial and temporal modeling completely, chunk-level features extract appearance and dynamics at the same time by creating a hierarchical representations of spatio-temporal data which leads to extracting more subject-related information. These features aggregate frames into one signature, then apply a deep neural network to extract both spatial and temporal information at the same time.
People use different strategies to encode chunk-level features: \emph{CNN-based architectures} and \emph{Attention}.

\textbf{CNN-based Standard Architectures.}
3D CNN encode a chunk of frames into one signature and has the kernel size of s×s×d which s and d refers to the kernel spatial size and the number of frames in one signature, respectively. In \cite{kong2017deep, kong2018adversarial,Yan_2019_CVPR}, a video was split into multiple segments, each of which is fed to a 3D CNN for feature extraction. This 3D CNN starts by focusing on the spatial relations for the first frames and then learns temporal dynamics in the following frames \cite{tran2015learning}. In \cite{yan2022gl}, two 3D CNN modules are utilized to encode both long and short temporal dependencies by taking chunks with different sizes.

Although 3D CNNs seem like a natural algorithm for modeling video, they have some drawbacks. First, they require a large number of parameters due to adding temporal dimension which leads to leveraging shallow architectures. In \cite{carreira2017quo}, a new model was introduced which used a 3D CNN with pre-trained Inception-V1 as a backbone. To bootstrap from the pre-trained ImageNet models \cite{5206848}, the weights of the 2D kernels are repeated along the time dimension. However, the fixed geometric structures of 3D convolution limits the learning capacity of the 3D networks. 

Second, chunk-level features learn temporal abstraction of high-level semantics directly from videos, however they are not suited for specific tasks that require granularity in time. In some applications, one may need to precisely predict dynamics in each frame. In this case, chunk level features loose granularity and cannot perform well. 
For instance, the temporal length of an input video is decreased by a factor of 8 in layers from conv1a to conv5b in C3D architecture \cite{tran2015learning}. This conforms that local information are lost passing multiple convolutions. To address this issue, in \cite{Shou_2017_CVPR}, a convolution and deconvolution approach is proposed which downsamples and upsamples in space and time, simultaneously. 

Last but not least, while 3D CNNs are well suited for capturing global coarse motions, they are limited in modeling finer temporal relations in a local spatio-temporal window. To address this, people use attention mechanism to exploit the temporal discriminative information in a chunk of frames.

\textbf{Attention Mechanism.} Chunk-level features usually learn the temporal domain by equally treating the consecutive frames in a chunk, while different frames might convey different contributions to the related task. Similar to the frame-level features, there are several works that explore temporal attention in chunk-level features \cite{li2020spatio,kim2021weakly}. In \cite{li2020spatio}, attention is learned at channel level by modeling the differences among the temporal channels in 3D CNNs.

\emph{Summary.} Chunk level features are suitable for capturing structural co-occurrence. They connect both spatial and temporal domains and are suitable for learning dynamics that differ in the order of their micro-movements. However, compared with frame-level features they are less suitable for tasks that require fine granularity in time. As shown in table~\ref{tab:chunk-level}, 3D CNNs can capture coarse level temporal dependencies in one signature. But they require a large number of parameters and cannot encode fine details in a local window. Attention is used to exploit informative features in both fine and coarse level and handle temporal scale variance. However, it adds extra weights to the model and is computationally expensive.

\section{Applying Deep Features in Video Analysis Tasks}
\label{Sec:Applications}

After discussing different feature modeling strategies, we compare their usages on different applications. 
We use two applications, namely, \emph{action recognition} and \emph{video object segmentation}, to analyze these features' behaviors under different circumstances. 
\emph{Action recognition} aims to recognize specific actions happened in a video and output one (or several, if there are multiple actions) global labels.
\emph{Video object segmentation} aims to identify and segment in pixel-level the objects of interest from background in every frame of a video. 

In both of these two applications, free-form deformations of objects are common in both spatial and temporal dimensions. 
And sometimes, to achieve better real-timer efficiency, temporal (or spatial) sampling is intentionally made sparse. 
Therefore, extracting reliable and powerful features plays a critical role and often directly dictates the final performance.

\subsection{Action Recognition} 
A key challenge in action recognition is how to learn a feature that captures the relevant spatial and motion cues with a descriptive yet compact representation. 
In the spatial dimension, some studies adopted dense features while some others used sparse features. Likewise, in the temporal dimension, some adopted frame-level features, while some others used chunk-level features. 
There is a lack of study that discusses the pros and cons of adopting these features under different circumstances. Hence, here we discuss and analyze each type of features, their challenges, limitations, and possible solutions as shown in table \ref{Tab:action}.

\begin{figure}[h!tb]
\centering
\includegraphics[width=130mm,scale=0.1]{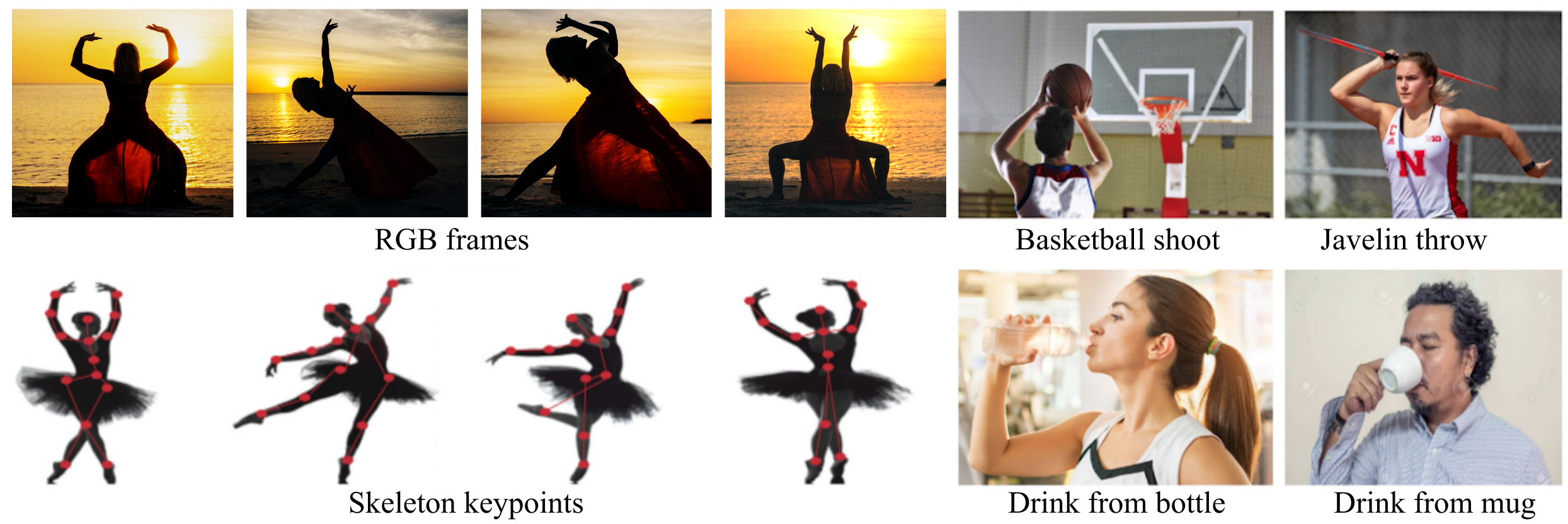}
\caption{\label{AR spatial} Dense (RGB frames) and sparse (skeleton keypoints) features in Action Recognition. 
Dense features may include background information; while sparse features encode mainly essential object structure. 
The usefulness of background varies:  it can either distract (e.g., in dancing images) or assist (e.g., activities in the right two columns) recognition.  RGB images in the upper left four columns are from \cite{pexels}; we put them in a sequence of frames. Skeleton keypoints in the lower left four columns are from \cite{lv2022complexity}.  Images in the right two columns are from UCF101 dataset \cite{soomro2012ucf101}.}
\end{figure}

\begin{table}[h!tb]
\centering
\footnotesize
\caption{\label{Tab:action} Pros and cons of spatial and temporal features in action recognition. There are some abbreviations in the table: ap., info., bg, rep. means appearance, information, background, and representation respectively.}

\begin{tabular}{|l|l|l|l|l|}
\hline
Features & Pros & Cons & Solution & References \\
\hline
\multirow{4}{*}{Dense}& \multirow{4}{*} {contains ap. info.} & bg noise/redundant info. & additional info. & \cite{Wang_2017_ICCV,zhao2021learning}\\
 & & & attention & \\
 & & rep. bias & calibrated data & \cite{li2018resound,girdhar2017attentional}\\
 & & & attention & \\
\hline

\multirow{2}{*}{Sparse} & robust to view/bg change & low reliability & additional info. & \cite{cai2021jolo, duan2022revisiting} \\
 & low computation cost & lack of scalibility & heatmap rep. &  \cite{choutas2018potion, yan2019pa3d}\\
\hline
Frame & low computation cost & co-occurrence rep. & message passing & \cite{yang2021feedback, si2019attention}\\
\hline
\multirow{2}{*}{Chunk} & \multirow{2}{*}{co-occurrence rep.} &
low computation cost & disentangle kernels & \cite{tran2019video,wang2021tdn} \\
& & fix-length chunk & multi scale kernels & \cite{dai2022ms,Hussein_2019_CVPR}\\
\hline
\end{tabular}

\end{table}

\textbf{Dense Features.}
\label{dense}
Due to the adaptability and availability of RGB video frames, such dense pixel-level representations are widely adopted for action recognition~\cite{ullah2021conflux,feichtenhofer2017spatiotemporal,xie2018rethinking}. 

\emph{Pros.} Dense features contain appearance information which is useful in recognizing actions in different scenes. As CNNs have shown their strong ability in capturing dense spatial features, majority of studies use either 2D or 3D CNNs to extract spatial semantics in video frames. Thanks to CNNs, modeling Dense features are straightforward compared to sparse features, however they have their task-related limitations.

\emph{Cons.} There are several challenges in using Dense features. First, they may contain background noise and redundant information which undermines robustness of action representation learning. Another limitation of dense features is ``representation bias" which refers to recognizing actions based on object/background detection. The network may predict correct results based on scene context rather than human action. Some actions might be easier to be predicted using the background and context, like a ‘basketball shoot’ vs a ‘throw’; some others might require paying close attention to objects being interacted by the human, like in the case of ‘drinking from mug’ vs ‘drinking from water bottle’ as shown in Fig. \ref{AR spatial}. It is noted to mention that representation bias is different from background noise. In background noise, the representation for each class of action differs with the change of the scene, while representation bias means getting help from the discriminative objects in the scene to recognize the action.  While some studies in action recognition consider representation bias undesired \cite{choi2019can,li2018resound}, it may be useful in some scenarios or similar tasks \cite{girdhar2017attentional}. The reason is that modeling human actions often requires understanding the people and objects around them. For example, recognizing ``listening to others" is not possible unless the model knows about the existence of another person in the scene saying something. The main concern with representation bias is that if the dataset is not generalized enough, it can undermine the performance of the model.

\emph{Solutions.}
To alleviate the background noise and redundant, several researches have augmented additional visual information to guide network from distraction. Some approaches adopted depth information to overcome the illumination and viewpoint variations \cite{qin2020dtmmn, Wang_2017_ICCV, Wang_2017_CVPR}, while others leverage skeleton data in the form of local attention to assist network in capturing the most representative body postures \cite{zhao2021learning,luvizon2020multi}.

 Solutions to overcome representation bias include collecting well-calibrated datasets \cite{li2018resound}, or using an attention mechanism to help the model focus on distinguishable parts of action \cite{sharma2015action}. Attention networks add a dimension of interpretability by capturing where the network is focusing when modeling actions. In \cite{sharma2015action}, the CNN produces a feature cube for each video input and predicts a softmax over locations and the label classes which determines the probability with which the model believes the corresponding region in the input frame is important. In \cite{girdhar2017attentional}, attention maps are produced to focus computation on specific parts of the input. The weighted attention pooling layer is plugged in as a replacement for a pooling operation in a fully convolutional network. In \cite{zheng2020global}, a three-stream architecture is proposed which includes two attention streams and a global pooling stream. A shared ResNet is used to extract spatial features for all three streams. Each attention layer employs a fusion layer to combine global and local information and produces composite features. Furthermore, global-attention regularization is proposed to guide two attention streams to better model dynamics of composite features with the reference to the global information.

\textbf{Sparse Features.}
Sparse features, particularly skeletons, are very popular in action recognition 
due to their action-focusing nature and compactness. Several studies use human skeleton information as a sequence of joint coordinate lists \cite{yan2018spatial,zhang2017view,li2018skeleton} where the coordinates are extracted by pose estimators. 

While using CNN networks to process RGB frames is straightforward, in skeleton-based action modeling, network is faced with the challenge of arranging skeleton features. Earlier methods \cite{zhang2017view,li2018skeleton} simply use the keypoint coordinates to generate a sequence of feature vectors. The issue with this method is that it focuses on modeling the information in the time domain and doesn't explore the spatial relations between body joints. Other approaches arranged skeleton data as a pseudo-image \cite{ke2017new,kim2017interpretable,li2017skeletonnnn}. However, recent works have shown that graph networks can efficiently model non-Euclidean data like human skeletons. Performance results from table \ref{Tab:arrangement} shows the superior performance of arranging skeleton data in a graph structure.

\begin{table}[h!tb]
\centering
\caption{\label{Tab:arrangement} Comparison of skeleton-based action recognition performance for NTU RGB-D \cite{shahroudy2016ntu} dataset with different feature arrangements. mAP refers to mean average precision.}
\begin{tabular}{|p{0.12\textwidth} | p{0.06\textwidth} | p{0.06\textwidth}|p{0.06\textwidth} | p{0.1\textwidth} | p{0.1\textwidth}|p{0.06\textwidth} | p{0.06\textwidth} | p{0.06\textwidth}|}
  \hline
  model & \cite{du2015hierarchical} & \cite{shahroudy2016ntu} & \cite{liu2016spatio} & \cite{kim2017interpretable} & \cite{ke2017new} & \cite{yan2018spatial} & \cite{wen2019graph} & \cite{li2019actional} \\
  \hline
  arrangement & vector & vector & vector & pseudo-image & pseudo-image & graph & \textbf{graph} & \textbf{graph} \\
  \hline
  mAP & 59.1\% & 62.9\% & 69.2\% & 74.3\% & 79.6\% & 81.5\% & \textbf{84.2}\% & \textbf{86.8}\% \\
  \hline
\end{tabular}

\end{table}

With introduction of ST-GCN in \cite{yan2018spatial}, spatio-temporal graph convolutions became a research hotspot. The core of this approach is to integrate temporal module in the spatial GCN. Several variants of ST-GCN are proposed \cite{cheng2020skeleton, wu2019spatial,cheng2020decoupling} to improve the network capacity and computation consumption of the network. However, the main limitation of ST-GCN is that it ignores the semantic connections among intra-frame joints by using a fixed graph structure. Recognizing action lies in looking beyond the local joint connectivity as learning not only happens in spatially connected joints, but also in the potential dependence of disconnected joints. For example, in ``walking" there is a high correlation between arms and legs while they are spatially apart. This achieves by extracting multi-scale features and long-range temporal dependencies, as joints that are spatially apart can also have strong correlations \cite{Liu_2020_CVPR}. In this regard, some techniques have been adopted to enhance the flexibility of GCNs. Attempts from using adaptive learning graph structure in \cite{shi2019skeleton,shi2020skeleton} to designing a graph-based search space to explore spatio-temporal connections \cite{peng2020learning} has been made. In \cite{dai2019tan}, dilated convolutions \cite{yu2015multi} are adopted to increase receptive field size and capture multi-scale context without increasing model complexity. In \cite{Liu_2020_CVPR} this issue is addressed by performing graph convolutions with higher-order polynomials of the skeleton adjacency matrix which increases the receptive field of graph convolutions. Attention mechanisms are also adopted to improve the ability of extracting high-level joints. In \cite{shi2020skeleton} a spatio-temporal channel attention module is embedded in each layer of the GCN, which enables model to focus on the discriminative details of joints, frames and channels in action recognition.     

\emph{Pros.} When sparse features are used, since only pose information is included, they contain high-level semantic information in a small amount of data and are more robust in dynamic circumstances \cite{shi2020decoupled,iqbal2017pose}. As skeleton data does not contain color information, it is not affected by the limitations of RGB frames \cite{si2019attention}, and can provide a stable low-frequency representation of human actions.

\emph{Cons.} There are some challenges in modeling sparse feature representation. A first limitation is the ``reliability" which means the recognition ability of sparse features is mainly affected by the distribution shift of coordinates. As joint coordinates are produced by a pose estimator, applying a different pose estimation algorithm may lead to a small perturbation of coordinates which causes different predictions \cite{zhu2019robust}. Also, local subtle motion patterns are lost in the process of pose estimation. Therefore, sparse nature of skeleton sequences is sometimes not informative enough for describing subtle actions like ``reading", ``writing", and ``shaking head". 

Another challenge is the lack of scalability of sparse-features. As sparse features are defined for every human separately. For example, each joint of human skeleton is defined as a node per person, the complexity of network linearly increases with increasing the number of persons, which limits its applicability in multi-person or group activity recognition. 

\emph{Solutions.} To overcome reliability issues, many approaches take advantage of multi-modal visual resources including RGB frames, depth maps and joint heatmaps to compensate for the lack of information in local and global domain \cite{cai2021jolo, duan2022revisiting}. People use other forms of representations using heatmap volumes to show skeletons to alleviate the scalibility issue of sparse features \cite{choutas2018potion, yan2019pa3d}.

\textbf{Frame-level Features.}
Some approaches extract the motion cues between adjacent frames and learn frame-level temporal dependencies in a sequential signature. In action recognition, modeling both short-range and long-range motions is sometimes required. 
(1) To this end, some earlier methods firstly extract hand-crafted optical flow \cite{simonyan2014two,wang2016temporal}, then use a 2D CNN-based two-stream framework to process optical flow and RGB frames in each stream separately. These lines of works have several drawbacks: First, computing optical flow is time-consuming and storage demanding. Furthermore, the training of spatial and temporal features is separated, and the fusion of two streams is performed only at the late layers. Several following works had improved this framework by using different mid-level links to fuse the features of two separated streams \cite{feichtenhofer2017spatiotemporal,feichtenhofer2016convolutional}. However, these methods still require additional time and storage costs for computation of optical flow. (2) Another line of work aggregates temporal information by sequence learning \cite{donahue2015long,li2018videolstm}. The majority of these methods treat each frame, or point in time, with equal weight, but not all parts of the video are equally important and thus it is also key that we develop feature extraction methods that can determine where to extract features from. First, non-uniformly extracting features efficiently from only informative temporal episodes is challenging as it is required to look at the whole video to determine which parts are informative. Some recent work \cite{korbar2019scsampler, wu2019multi, meng2020ar} have proved that a recognition system can benefit from selecting the informative frames rather than simply taking the uniformly sampled frames as inputs. However, these systems treat the frame selection and feature extraction as two separate stages and thus the frame selection can not benefit from the later feature extraction thus reducing the descriptive power of the network and adding redundancy in the two stages. 
In order to tackle this challenge, in \cite{li2022nuta}, a two-branch architecture is suggested that maintains both uniformly frame-level features and non-uniformly chunk-level features in an end to end manner. To produce non-uniformly features, a temporal map is used that non-uniformly projects temporal instances to a smaller subset by using self-attention-like module. This component is proposed to only sample the most informative frames across time. 
(3) Another recent line of work adopts video transformers that apply self-attention to spatial-temporal features. Representative networks include TimeSformer \cite{bertasius2021space}, ViViT \cite{arnab2021vivit}, Mformer \cite{patrick2021keeping} and MViT \cite{fan2021multiscale}. Combining 2D backbones and Transformers, VTN \cite{neimark2021video} and CARL \cite{chen2022frame} can efficiently process long video sequences. However, these networks are designed to process a batch of frames at once on video tasks which requires large computing memory. In \cite{yang2022recurring}, a recursive mechanism is deployed to process the videos frame by frame and consume less GPU memory. 

\emph{Pros.} Generally, standard architectures of frame-level features have lower computational costs compared to chunk-level features. Also, frame-level features are more concise when aggregating local frame-level features are required for a global compact representations. 

\emph{Cons.} The main challenge in extracting frame-level features is ``co-occurrence representation" learning which in action recognition refers to when a model needs to learn a set of human actions with a specific set of spatial features at certain times. For example, in the action of "touching back", model needs to focus first on the hand and then pay attention to the back \cite{li2020global}. In many of existing approaches, a temporal module and a spatial module are designed separately. Their typical approach is to use a convolutional network to extract spatial relations in each frame, then use a 1D convolution \cite{Shi_2019_CVPR,shi2019skeleton,li2019actional} or LSTM \cite{Li_2018_CVPR,li2019spatio} to model temporal dependencies. However, such decoupled design restricts the direct information flow across space-time for capturing complex regional spatial-temporal joint dependencies. 

\emph{Solutions.} To help model better learn co-occurrence representation, message passing and cross connection strategy is adopted to avoid stacking multiple spatio-temporal modules and transfer information. In \cite{yang2021feedback}, the feedback connection was integrated into GCN to transfer the high-level semantic features to the low-level layer, and gradually transmitted the temporal information to build the global spatio-temporal action recognition model. In \cite{tran2017two}, attentions provided from temporal-stream is used to help spatial stream by cross-link layers. In \cite{li2020global}, a coordinate system conversion and spatio-temporal-unit feature enhancement is proposed to perform co-occurrence learning. In \cite{si2019attention}, each joint coordinate is transformed into a spatial feature with a linear layer. Then data is augmented with the spatial feature and the difference between spatial features between consecutive frames. Then a shared LSTM and three layers of graph convolutional LSTM are applied to model co-occurrence representation learning between joints.

\textbf{Chunk-level Features.}
\label{chunk}
Another type of approaches is to extract chunk-level temporal features in a global signature. In this case, the chunk is defined as multi-dimensional time series of dense/sparse features. Thanks to the ability of 3D CNNs to implicitly model motion information along with the semantic features, this line of works has seen significant advances in recent years. The first work in this line was C3D \cite{tran2015learning}, which proposed using 3D convolutions to jointly model the spatial and temporal features in a global signature. To use pre-trained 2D CNNs, in \cite{carreira2017quo}, I3D was proposed that inflates the pre-trained 2D convolutions to 3D ones. 

\emph{Pros.} Generally, chunk-level features benefit from the co-occurrence representation learning as there is a link between temporal and spatial channels. There networks are potentially more effective in learning fine detailed actions such as ``sitting" and ``standing up". 

\emph{Cons.}  While so many attempts have been done in capturing motion using chunk-level features, most of approaches often lack specific consideration in the temporal dimension. Therefore, designing an effective temporal module of high motion modeling power and low computational consumption is still a challenging problem. First, the 3D networks require a substantial amount of computation and time. Also, compared to 2D kernels, 3D convolutions have to reduce the spatial resolution to decrease memory consumption, which may lead to the loss of finer details. 

Moreover, temporal features are typically extracted from a fixed-length clip instead of a length-adaptive clip which is not suited for different visual tempos. Visual tempo defines the speed of an action, meaning how fast and slow an action is performed at the temporal scale which in some cases is crucial for recognizing actions e.x. walking, jogging and running.

\emph{Solutions.} To decrease the heavy computations of 3D CNNs, some works proposed to factorize the 3D convolution with a 2D spatial convolution and a 1D temporal convolution \cite{lin2019tsm,he2019stnet,tran2019video} or a mixed up of 2D
CNN and 3D CNN \cite{xie2018rethinking,zolfaghari2018eco}. In \cite{tran2019video}, a group convolution is used to disentangle channel interactions and spatio-temporal interactions, or use separated channel groups to encode both spatial and spatio-temporal interactions in parallel with 2D and 3D convolution \cite{luo2019grouped}. While all these existing approaches are designed to deal with the curse of dimension, there is a lack of data dependent decision to adaptively guide features through different routs in the network. In \cite{wang2021tdn}, two temporal difference module is proposed which computes multi-scale and bidirectional motion information between frames and chunks. In \cite{sudhakaran2020gate}, features are selectively routed through temporal dimension and are combined with each other without any computational overhead. 

Some people use multi-scale convolutional kernels to cover various visual tempos. In \cite{yang2019asymmetric}, multi-scale convolutional features are incorporated in asymmetric 3D convoultions to improve temporal feature learning capacity. In \cite{Hussein_2019_CVPR}, Timeception layer is designed which temporally convolves each chunk using multi-scale temporal convolution module to tolerate a variety of temporal extents in a complex action. Some people design a level-specific network frame pyramid to handle the variance of visual tempos \cite{feichtenhofer2019slowfast, zhang2018dynamic}. In \cite{dai2022ms}, a multi-scale transformer is proposed which is built on top of temporal segments using 3D convolutions in a token-based architecture to promote multiple temporal scales of tokens. Having different scales allows model to capture both fine-grained and composite actions across time. In \cite{truong2022direcformer}, a direct attention mechanism is incorporated in transformers to exploit the direction of attention across frames and correct the incorrectly-ordered frames to the right ones and provide an accurate prediction.

\begin{table}[h]
\centering
\caption{\label{Tab:quant} Comparison of different action recognition performance for NTU RGB-D \cite{shahroudy2016ntu} dataset. CS and CV refer to cross subject (various human subjects) and cross view (various camera views) split of the dataset.}

\begin{tabular}{|p{0.1\textwidth} |p{0.2\textwidth}|p{0.15\textwidth} |p{0.18\textwidth}|p{0.18\textwidth}|}
\hline
  Model & Spatial & Temporal & CS accuracy & CV accuracy\\ \hline
  \cite{baradel2018glimpse} & Dense & Frame & 86.6\% & 93.2\% \\
  \cite{zhu2018action} & Dense & Chunk & 94.3\% & 97.2\% \\
  \cite{caetano2019skelemotion} & Sparse & Chunk & 76.5\% & 84.7\% \\
  \cite{thakkar2018part} & Sparse & Chunk & 87.5\% & 93.2\% \\
  \cite{de2020infrared} & Dense + Sparse & Chunk & 91.8\% & 94.9\% \\
  \cite{liu2018recognizing} & Dense + Sparse & Frame & 91.7\% & 95.2\% \\
  \cite{su2020msaf} & Dense + Sparse & Frame & 92.2\% & - \\
  \cite{bruce2021multimodal} & Dense + Sparse & Chunk & 92.5\% & 97.4\% \\
  \cite{duan2022revisiting} & Dense + Sparse & Chunk & 97.0\% & 99.6\% \\
  \cite{bruce2022mmnet} & Dense + Sparse & Chunk & 96.0\%	& 98.8\% \\\hline
\end{tabular}

\end{table}

\textbf{Summary.} We summarized pros and cons of different features and the possible solution in table \ref{Tab:action}. Dense features have the advantage of using appearance info in action recognition while they suffer from background noise and representation bias. Possible solutions for background noise and representation bias include augmenting additional information to the input and well calibrated dataset and attention mechanism, respectively. Sparse features are more robust against background noise and have lower computational cost compared to dense features, while they suffer from lack of reliability and scalibility. Possible solutions for their drawbacks could be augmenting additional information and visual sources. Researches have shown in spatial domain, using multi-modal inputs, particularly accompanied with attention mechanism are very helpful in understanding human actions, as humans' brain adopt all visual inputs to recognize an action. The quantitative results are shown in table \ref{Tab:quant} on NTU RGB-D dataset confirm that using multi-modal spatial features outperforms single-modal approaches.

In temporal domain, using frame-level features has the advantage of lower computational cost, compared to chunk-level features. However, decoupling spatial and temporal dimensions restricts co-occurrence representation learning in distinguishing complex actions.  Some studies alleviate this problem by message passing techniques and cross-link connections. On the contrary, chunk-level features allow establish of connections and links between temporal and spatial dimensions to learn order and co-occurrence of micro-actions. However, typically these models take fixed-length instead of adaptive-length chunks. Some works address this issue by using frame pyramids and multi-scale convolutions. Another drawback of chunk-level features is their high computational costs which could be alleviated by disentangling convolutions in different layers of network. To conclude, as shown in table \ref{Tab:action}, in presence of sparse features, using chunk-level approaches outperforms frame-wise methods due to reducing number of parameters, increasing depth of network and co-occurrence representation learning. 

\subsection{Video Object Segmentation}
Video object segmentation (VOS) is a video processing technique.
The goal of VOS is to segment pixel-level masks of foreground objects in every frame of a given video.
VOS has attracted extensive attention these years because it can be applied to diverse fields in computer vision. 
Recent VOS research can be divided into two sub-tasks: semi-supervised and unsupervised.
The semi-supervised VOS aims to re-locate and segment one or more objects that are given in the first frame of a video in pixel-level masks.
The unsupervised VOS aims to automatically segment the object of interest from the background, usually, the most salient object(s) will be segmented. 
Generally, the input to the video object segmentation is a sequence of color frames in RGB format.
Feature modeling is the first stage of these video object segmentation pipelines.
In the following, we discuss pros and cons of different features in VOS.

\begin{table}
    \centering
    \caption{Pros and Cons of spatial and temporal features in the video object segmentation (VOS) task. There are some abbreviations in the table: ap. and info. means appearance and information, respectively.}
    \begin{tabular}{|p{0.08\textwidth} |p{0.22\textwidth} |p{0.23\textwidth} |p{0.2\textwidth} |p{0.1\textwidth} |}
    \hline
    Features & Pros & Cons & Solution & References \\
    \hline
    Dense & contains rich ap. info. & weak in occlusion-handling. & occlusion-aware encoders & \cite{ke2021deep, ke2021occlusion}\\
    \hline
    Sparse & fast \& low comp. & less accuracy & hybrid algorithm & \cite{chen2020state, sun2020fast} \\
    \hline
    \multirow{2}{*}{Frame} & low computation cost & error accumulation & dynamic feature & \cite{liang2020video, li2022recurrent}\\
    & low latency & past frame management & space management & \\
    \hline
    Chunk & multi-modal modeling &
    large computation cost & knowledge distillation & \cite{wu2022language, wu2022multi} \\
    \hline
    \end{tabular}
    \label{tbl:vos-features}

\end{table}


\textbf{Dense Features.} Dense features are widely used because of the neutrality of VOS which is supposed to estimate the pixel-level object masks.
These methods often apply a pre-trained CNN encoder to extract dense feature maps from each frame. 
Generally, 2D CNN encoder is widely-used to extract feature maps~\cite{liu2021sg, athar2022hodor, park2022per, lin2022swem}.
The CNN encoder is often pretrained on ImageNet~\cite{deng2009imagenet} and fine-tuned on the video object segmentation dataset.
After extracting the dense feature maps, these methods utilize a transformer \cite{vaswani2017attention} to encode the feature maps into two \emph{keys} and \emph{values}, where \emph{keys} contains the semantic code of the object and \emph{values} contains the detailed appearance information. 
The encoded feature maps can be used for matching and information retrieval. Besides the appearance feature maps, Zhang et al.~\cite{zhang2022perceptual} introduced perceptual consistency to aid with predicting the pixel-wise correctness of the segmentation on an unlabeled frame.

\emph{Pros.} RGB frame provides rich information about the appearances textcolor{green}{which is crucial in VOS}. With the help of GPU parallel computing and large-scale pretrained model,  dense features extraction from RGB frame becomes standard processing in video object segmentation nowadays.

\emph{Cons.}
Although dense feature extraction has been widely studied, They may contain background noise and irrelevant information. Particularly, occlusion-handling is a main challenge of using dense features in VOS. 

\emph{Solutions.} 
In dealing with occlusions, its imperative to recover occluded parts by corresponding shape and appearance in motion rather than irrelevant background. For this purpose, Occlusion-aware feature modeling ~\cite{ke2021occlusion} is proposed which uses shape completion and flow completion modules to inpaint invisible parts intelligently. In \cite{ke2021deep}, a pipeline is proposed that used GCN which allows propagation of non-local information across pixels despite the presence of occluding regions.

\textbf{Sparse Features.}
Compared with the dense feature extraction, sparse feature modeling discards the irrelevant information from input. In VOS, short tracks or tracklets in a frame are considered as sparse features. In ~\cite{chen2020state}, a State-Aware Tracker (SAT) is proposed that takes advantage of the inter-frame consistency and deal with each target object as a tracklet. Because the irrelevant background is discard, they achieve real-time video object segmentation on 39 FPS which is faster than traditional dense feature methods. In ~\cite{sun2020fast}, the video object tracking module is adopted to first locate the object region from the background. Then they segment the object masks from the small object region.

\emph{Pros.}
The advantages of the sparse feature modeling are two fold.
Firstly, it uses pre-processing or prior knowledge to clean the irrelevant information from input, which makes the pipeline more robust to the noises from the background.
Secondly, using sparse features can accelerate the speed of segmentation, which gives the users an option to choose a trade-off between efficiency and accuracy.

\emph{Cons.}
Sparse feature modeling in VOS relies on the data pre-processing. It is inevitable that some important data from the object is mistakenly discard in this process. Also, sparse features lack the appearance/color information which is helpful in VOS.

\emph{Solution.}
Compared with dense feature modeling, sparse feature modeling has lower accuracy but better runtime performance. Recently, \cite{Wu2022EfficientVIS} used tracklet query and tracklet proposal that combines RoI features and dense frame features by the vision transformer. It achieved state-of-the-art performance in YouTube-VIS 2019 val set~\cite{yang2019video}.

\begin{table}[ht]
\centering
\caption{Comparison of different spatial feature performance for DAVIS-2016 and DAVIS-2017 datasets~\cite{DAVIS2017}.  $\mathcal{J}$, $\mathcal{F}$ and $\mathcal{J\&F}$ score represent region similarity, contour accuracy and the average value of region similarity and contour accuracy, respectively. All models use frame-level features in temporal dimension. }
\label{Tab:quant-vos-davis}
\begin{tabular}{|p{0.16\textwidth} |p{0.07\textwidth} 
|p{0.05\textwidth} |p{0.05\textwidth} |p{0.08\textwidth} |p{0.07\textwidth} |p{0.07\textwidth} |p{0.08\textwidth} |p{0.06\textwidth} |}
    \hline
    \multirow{2}{*}{Model}& \multirow{2}{*}{Spatial} & \multicolumn{3}{c}{DAVIS val 16} & \multicolumn{3}{c}{DAVIS val 17} \\
    & & $\mathcal{J} (\%)$ & $\mathcal{F} (\%)$ & $\mathcal{J\&F} (\%)$ & $\mathcal{J} (\%)$ & $\mathcal{F} (\%)$ & $\mathcal{J\&F} (\%)$ & FPS \\
      \hline
      FEELVOS~\cite{voigtlaender2019feelvos} & Dense & 81.1 & 82.2 & 81.7 & 69.1 & 74.0 & 71.5& 2.2\\\footnotesize
      STM~\cite{oh2019video}                & Dense  & 84.8 & 88.1 & 86.5 & 69.2 & 74.0 & 71.6 & 6.3\\
      AGAME~\cite{johnander2019generative} & Dense & 82.0 & - & - & 67.2 & 72.7 & 70.0 & 14.3\\
      AGSS-VOS~\cite{lin2019agss}               & Dense & - & - & - &  63.4 & 69.8 & 66.6 & 10.0 \\
      AFB-URR~\cite{liang2020video} & Dense & - & - & - & 73.0 & 76.1 & 74.6 & 4.0\\
      FRTM~\cite{robinson2020learning}       & Dense & - & - & 81.7 & 66.4 & 71.2 & 68.8 & 21.9 \\
      LCM~\cite{hu2021learning}             & Dense & - & - & - & 73.1 & 77.2 & 75.2 & 8.5 \\
      RMNet~\cite{xie2021efficient}         & Dense & 80.6 & 82.3 & 81.5 & 72.8 & 77.2 & 75.0 & 11.9 \\
      SWEM~\cite{lin2022swem}               & Dense & \textbf{87.3} & \textbf{89.0} & \textbf{88.1} & \textbf{74.5} & \textbf{79.8} & \textbf{77.2} & 36\\
      BMVOS~\cite{cho2022pixel}             & Dense & 82.9 & 81.4 &  82.2 & 70.7 & 74.7 & 72.7 & 45.9\\
      FTM~\cite{sun2020fast}        & Sparse & 77.5 & - & 78.9 & 69.1 & - & 70.6 & 11.1 \\
      SAT~\cite{chen2020state}      & Sparse & 82.6 & 83.6 & 83.1 & 68.6 & 76.0 & 72.3 & 39.0 \\
      SAT-Fast~\cite{chen2020state} & Sparse & - & - & - & 65.4 & 73.6 & 69.5 & \textbf{60.0} \\
    \hline
\end{tabular}
\end{table}

\begin{table}[t!hb]
\centering
\caption{Comparison of different spatial feature performances for Youtube-VOS~\cite{xu2018youtube} dataset. $\mathcal{J}$ and $\mathcal{F}$ represent region similarity, contour accuracy and seen and unseen tags indicate seen and unseen objects from the training dataset separately. $\mathcal{G}$ measures the overall score of the segmentation accuracy. FPS
metric reports the runtime speed in frames per second.}
\label{Tab:quant-vos-youtube}
\begin{tabular}{|p{0.13\textwidth} |p{0.07\textwidth} 
|p{0.09\textwidth} |p{0.12\textwidth} |p{0.09\textwidth} |p{0.12\textwidth} |p{0.05\textwidth} |p{0.05\textwidth}|}
    \hline
    Model & Spatial & $\mathcal{J}_{seen} (\%)$ & $\mathcal{J}_{unseen} (\%)$ & $\mathcal{F}_{seen} (\%)$ & $\mathcal{F}_{unseen} (\%)$ & $\mathcal{G}$ & FPS \\
      \hline
      STM~\cite{oh2019video}        & Dense & 79.7 & 84.2 & 72.8 & 80.9 & 79.4 & - \\
      RVOS~\cite{ventura2019rvos}   & Dense & 63.6 & 45.5 & 67.2 & 51.0 & 56.8 & 22.7 \\
      AGSS-VOS~\cite{lin2019agss}   & Dense & 71.3 & 65.5 & 76.2 & 73.1 & 71.3 & 12.5 \\
      AFB-URR~\cite{liang2020video} & Dense & 78.8 & 74.1 & \textbf{83.1} & 82.6 & 79.6 & 3.3 \\
      FRTM~\cite{robinson2020learning}    & Dense & 72.3 & 65.9 & 76.2 & 74.1 & 72.1 & -\\
      LCM~\cite{hu2021learning}     & Dense & 82.2 & 75.7 & 86.7 & 83.4 & 82.0 & - \\
      RMNet~\cite{xie2021efficient} & Dense & 82.1 & 85.7 & 75.7 & 82.4 & 81.5 & - \\
      SWEM~\cite{lin2022swem}       & Dense & \textbf{82.4} & \textbf{86.9} & 77.1 & \textbf{85.0} & \textbf{82.8} & - \\
      BMVOS~\cite{cho2022pixel}     & Dense & 73.5 & 68.5 & 77.4 & 76.0 & 73.9 & 28.0 \\
      SAT~\cite{chen2020state}      & Sparse & 67.1 & 55.3 & 70.2 & 61.7 & 63.6 & \textbf{39.0} \\
    \hline
\end{tabular}

\end{table}

\textbf{Frame-level Features.} Several VOS methods predict segmentation masks frame by frame. 
They utilize previous frames information to model the current object appearance and motion. Appearance similarity between frames is widely-used to segment objects from background~\cite{wang2019ranet, oh2019fast, xu2019spatiotemporal, johnander2019generative, zhang2022perceptual, zhou2022slot, li2022recurrent, athar2022hodor}. 
The intuition behind these papers is that  perceptually similar pixels are more likely to be in the same class. In frame-level features, it is essential to make full use of historical  frames in the videos. Some methods use space-time memory banks to store the embeddings every several frames \cite{cheng2021modular,hu2021learning}. Other methods fuse pixel embedding of the current frame and the memory bank \cite{liang2020video}.

In ~\cite{ventura2019rvos} a recurrent pipeline (RVOS) is proposed to keep the coherence of the segmented objects along time. Due to the RNN's memory capabilities, RVOS is recurrent in the spatial-temporal domain and can handle the instances matching at different frames. Recent papers used auxiliary temporal information including optical flow ~\cite{zhang2019fast, lin2019agss, azulay2022temporally} to aid VOS.

\emph{Pros.}
Most VOS methods follow the frame-level feature modeling because it usually has lower computational costs. These features are also suitable to handle streaming data such as online video or online meeting because it processes the video frame by frames. Using optical flow map as dense map is very common in the unsupervised video object segmentation task~\cite{ren2021reciprocal, zhang2021deep}, which provide the dense correspondence between similar pixels. 

\emph{Cons.}
Although frame-level feature modeling has lots of benefits, it brings some shortcomings and limitations. First of all, as information is processed frame by frame, past frames aid model in segmenting objects in the current and future frames. Managing past information is very challenging and requires lots of efforts in model design, particularly when the length of video increases. Additionally, the errors from past frames are accumulated and transferred during video processing.

\emph{Solutions.}
Recent research papers proposed adaptive memory bank scheme to maintain features from previous frames. In ~\cite{liang2020video} an exponential moving average (EMA) based scheme is proposed to update the record of historical information. This method helps model to better use past information. In ~\cite{li2022recurrent}, a learning based spatial-temporal aggregation model (SAM) is introduced to distill the frame-level features and automatically correct the accumulated errors.

\textbf{Chunk-level Features.}
Inspired by the human action recognition field, In ~\cite{huang2020fast}, a 3D convolutional neural network is used as a backbone to extract chunk-level features. Recent chunk-level feature modeling is used in the multimodal video segmentation task, which combines text reasoning, video understanding, instance segmentation and tracking. Recent papers first use a CNN encoder to extract feature maps from each frame \cite{qin2023coarse}, then combine them into the spatial-temporal features. In~\cite{botach2022end, wu2022language, wu2022multi}, a Transformer is used to model spatial-temporal features from videos.

\emph{Pros.}
The advantage of chunk-level features in VOS is that it models the global semantic information in consecutive frames, which is critical to multi-modal tasks.

\emph{Cons.}
Modeling chunk-level features often requires large memory and computation costs compared with frame-level features. Heavy computation requirement limits its application on mobile device.

\emph{Solutions.}
Knowledge distillation~\cite{lin2022knowledge} and neural network pruning/search~\cite{zhou2022training} are two active fields aim to relief the burden of neural network computing. They can help in finding a smaller network with lower network parameters but similar accuracy performance. These methods could be used to optimize the architecture of the chunk-level feature encoder to accelerate their computation.
In ~\cite{xu2022accelerating}, an acceleration framework is proposed based on video-compressed codec. 
For each chunk, it has three types of I-frames, P-frames, and B-frames. They utilize the sparsity of I-frames and motion vectors from P-/B-frames to accelerate the video object segmentation.

\begin{table}[t!hb]
\centering
\caption{Comparison of different models performance with chunk level features for Ref-Youtube-VOS~\cite{seo2020urvos} across different backbones and chunk sizes $\omega$. $\mathcal{J}$, $\mathcal{F}$ and $\mathcal{J\&F}$ score represent region similarity, contour accuracy and the average value of region similarity and contour accuracy, respectively.}
\label{Tab:quant-vos-ref-youtube}
\begin{tabular}{|p{0.3\textwidth}|p{0.2\textwidth}|p{0.1\textwidth}|p{0.1\textwidth}|p{0.12\textwidth}|}
    \hline
    Model & Backbone & $\mathcal{J} (\%)$ & $\mathcal{F} (\%)$ & 
    $\mathcal{J\&F} (\%)$\\
      \hline
      URVOS~\cite{seo2020urvos} & ResNet-50 & 45.3 & 49.2 & 47.2\\
      MLRL~\cite{wu2022multi} & ResNet-50 & 48.4 & 51.0 &  49.7\\
      MTTR ($\omega$=12)~\cite{botach2022end} & Video-Swin-T & 54.0 & 56.6 & 55.3 \\
      ReferFormer ($\omega$=5)~\cite{wu2022language} & Video-Swin-T &  54.8 & 57.3 & 56.0\\
      ReferFormer~\cite{wu2022language} & Video-Swin-B & \textbf{61.3} &\textbf{ 64.6} & \textbf{62.9}\\
    \hline
\end{tabular}
\end{table}

\begin{table}[t!hb]
\centering
\caption{Comparison of different spatial feature performance for A2D-Sentences~\cite{gavrilyuk2018actor}. $*^T$ indicates the backbone architecture is Video-Swin-T. $*^B$ indicates the backbone architecture is Video-Swin-B. The Precision @K measures the percentage of test samples that their whole IoU scores are higher than threshold K.}
\label{Tab:quant-vos-a2ds}
\begin{tabular}{|l|p{0.055\textwidth} | p{0.055\textwidth} | p{0.055\textwidth}| p{0.055\textwidth} | p{0.055\textwidth} | p{0.058\textwidth}|p{0.055\textwidth} | p{0.055\textwidth}|}
    \hline
    \multirow{2}{*}{Model} & \multicolumn{5}{c}{Precision} & \multicolumn{2}{c}{IoU} & \multirow{2}{*}{mAP}\\
    & P@0.5 & P@0.6 & P@0.7 & P@0.8 & P@0.9 & Overall & Mean &  \\
      \hline
      MTTR$^T$($\omega$=8)~\cite{botach2022end}  & 72.1 & 68.4 & 60.7 & 45.6 & 16.4 & 70.2 & 61.8 & 44.7 \\
      MTTR$^T$($\omega$=10)~\cite{botach2022end}  & 75.4 & 71.2 & 63.8 & 48.5 & 16.9 & 72.0 & 64.0 & 46.1 \\
      ReferFormer$^T$($\omega$=6)~\cite{wu2022language} & 76.0 & 72.2 & 65.4 & 49.8 & 17.9 & 72.3 & 64.1 & 48.6 \\
      ReferFormer$^B$($\omega$=5)~\cite{wu2022language} & \textbf{83.1} & \textbf{80.4} & \textbf{74.1} & \textbf{57.9} & \textbf{21.2} & \textbf{78.6} & \textbf{70.3} & \textbf{55.0} \\
    \hline
\end{tabular}
\end{table}

\textbf{Summary.}
We summarized pros and cons of features and the possible solution in table 10. Dense features Contain rich appearance information which is essential in VOS, but they are not robust against occlusion. Some approaches use occlusion-aware encoders to overcome this shortcoming. Sparse features disregard irrelevant information and have lower computation cost. The accuracy of sparse features are lower in VOS task. To overcome this issue, people use hybrid algorithms. Frame-level features have lower computational cost compared to frame-level features and are suitable for stream processing. However, the computation error is accumulated in these features while using past information. Also, managing past frames requires a lot of efforts. Some researches adopted dynamic memory to relive the error accumulation. Also they used exponential weight smoothing and learning based feature adaption to manage past frame information. Chunk-level features model more global information than frame-level features but are more computationally expensive. Different methods are adopted that use knowledge distillation to alleviate computation burden. 

To conclude as shown in Table~\ref{Tab:quant-vos-davis} and Table~\ref{Tab:quant-vos-youtube}, we compared the performance of different spatial feature models.
These comparisons are from the spatial aspect, they all belong to the frame-level features.
The DAVIS'17 datasets~\cite{DAVIS2017} benchmarks contain 60 videos for training and 30 videos for validation.
The YouTube-VOS~\cite{xu2018youtube} benchmark contains 3,471 videos for training and 507 for validation.
In both benchmarks, the video object segmentation task is to segment and track the arbitrary number of objects in each video.
The groundtruth masks for each object are provided in the first frame.
For DAVIS datasets, we followed the official evaluation metrics, region similarity $\mathcal{J}$ and contour accuracy $\mathcal{F}$. 
The $\mathcal{J\&F}$ score is the average value of region similarity and contour accuracy.
For YouTube-VOS dataset, we used similar metrics $\mathcal{J}$ and $\mathcal{F}$.
$seen$ and $unseen$ tags indicate seen and unseen objects from the training dataset separately.
$\mathcal{G}$ measures the overall score of the segmentation accuracy.
FPS metric reports the runtime speed in frames per second.
The runtime speed was measured on NVIDIA 2080Ti and NVIDIA V100 GPUs.
SWEM~\cite{lin2022swem} has the best accuracy performance in both DAVIS and YouTube-VOS datasets because it utilizes dense feature modeling, while SAT~\cite{chen2020state} achieves the best runtime speed because of sparse feature modeling.

To compare the chunk-level feature modeling, we chose the popular Ref-Youtube-VOS~\cite{seo2020urvos} and A2D-Sentences~\cite{gavrilyuk2018actor} benchmarks for comparisons.
The Ref-YouTube-VOS dataset covers 3,978 videos with around $15K$ language descriptions. 
We used similar metrics to measure the segmentation performance, region similarity $\mathcal{J}$, $\mathcal{F}$, and  $\mathcal{J\&F}$.
The A2D-Sentences dataset contains 3,782 videos and each video has 3-5 frames annotated with the pixel-level segmentation masks.
The model is evaluated with criteria of Precision $@K$, Overall IoU, Mean IoU and mAP.
The Precision $@K$ measures the percentage of test samples whole IoU scores are higher than threshold K. 
Following standard protocol, the thresholds are set as 0.5:0.1:0.9.
We compared the performance results across different network settings: (1) different backbones Video-Swin-T and Video-Swin-B from Video Swin Transformer \cite{liu2022video}, (2) different chunk sizes $\omega$.
From Table~\ref{Tab:quant-vos-ref-youtube} and Table~\ref{Tab:quant-vos-a2ds}, ReferFormer~\cite{wu2022language} achieve the best performance because it designed the cross-model feature pyramid network to extract multi-scale chunk-level features from the input.

\section{Conclusion and Future Work}\label{Sec:Conclusion}
We provided an extensive survey on recent studies on deep video representation learning. 
We provided a new taxonomy of these features and classify existing methods accordingly. 
We discussed and compared the effectiveness (robustness) of different types of features under scenarios with different types of noise.

\textbf{Challenges.}
Spatially dense features can encode rich contextual information but are more sensitive to background noise. Handling dense features in presence of intense occlusion and view variations is still challenging. In contrast, sparse features are more robust against background noise and illumination variance, but arranging sparse topologies in spatial dimension is still challenging. It is still an open question if the spatial relations should be defined based on the natural inherent relation of the features points or on the correlation of non-adjacent points throughout the video. For example, in human body skeleton, should the spatial relations between coordinates be defined based on natural relations of human body joints or the correlation of non-adjacent joints during the movement?
In frame-level features, lack of cross connections between temporal and spatial domains is a major drawback in capturing complex dynamics. In chunk-level features, improving model’s generalizability and high computational cost are the main challenges. 

\textbf{Future directions.} 
The drawbacks of either using sparse or dense features could be solved by using multi modal inputs to some extent. A future direction of these studies is on designing new methods for mapping between different modalities' feature space, learning effective representations from multiple data modalities, and understanding when and where the fusion should happen. Recent studies~\cite{Zolfaghari_2021_ICCV} 
demonstrated using proper multi-modals clearly improve video analysis performance.
While current attention methods have achieved progress in video representation learning, they often bring higher model complexity and suffer from heavier computational burden. Hence, many recent studies are on building more efficient attention models~\cite{hou2021coordinate}.

\paragraph{Data availability statement} All data supporting the findings of this study are available within the paper.

\section*{Declarations}
\textbf{Competing interests} We do not have any conflict of interest related to the manuscript.

\bibliography{sn-bibliography}

\end{document}